\documentclass{article}
\usepackage[preprint]{neurips_2026}

\usepackage[utf8]{inputenc}
\usepackage[T1]{fontenc}
\usepackage{hyperref}
\usepackage{url}
\usepackage{booktabs}
\usepackage{amsfonts}
\usepackage{amsmath}
\usepackage{amssymb}
\usepackage{graphicx}
\usepackage{xcolor}

\title{Your Prompt Is Not the Only Prompt: How Much Do LLMs Weight \\ Structured-Output Schema Descriptions?}

\author{
  Sin-Ying (Alina) Lin \\
  \texttt{Independent Researcher} \\
}

\begin{document}

\maketitle

\begin{abstract}
Structured output, in which a large language model populates a predefined JSON schema, has become a default mechanism for LLM-based data labeling and information extraction due to its robustness in enforcing output structure. However, it also introduces a second instruction channel such that each schema property can contain a description field. Practitioners therefore face a recurring choice: where should content such as category definitions and behavioral constraints live---in the system prompt or in schema descriptions? Ambiguity remains even in vendor API examples.

We tested this question on a single-field classification task with nonce category labels across ten model configurations from two vendors, moving identical definition text between the system prompt, user prompt, and schema description. When instructions are consistent, schema descriptions did not consistently outperform prompt-based placement, and for two models (GPT-4.1 and GPT-5.4 without reasoning), schema placement underperformed system prompts by 11--13 percentage points. However, schema descriptions are not inert metadata. When system prompts and schemas conflict, incorrect schema instructions caused accuracy drops of 5--45 percentage points: Claude Haiku 4.5 fell from 52.5\% to 7\%, indicating that schema instructions can override prompt instructions, while GPT-5.5 fell from 100\% to 73\%. Additionally, adding a required intermediate reasoning field before the label field inside the schema improved schema-only accuracy by 15--24 percentage points when headroom existed, exceeding system-prompt-only performance in every case tested. This held even for Claude Sonnet 4.6 at medium reasoning, where the model's own extended-thinking mode alone did not produce a comparable gain.

Our results de-mystify the role of the schema rather than confirm either side of that folk debate: it is neither inert metadata nor a channel with the same weight as the system prompt. How much it matters is a model-specific property to be measured, making instruction placement an empirical question. In practice, the system prompt remains a safe default for definition placement. The real discipline is keeping definitions as a single source of truth rather than letting prompt and schema drift apart. However, prompt placement may not be the biggest lever at all. The reasoning-field intervention produced 15--24 point gains whether or not a model showed a placement penalty to begin with, exceeding that penalty wherever one existed. Overall, the results indicate that practitioners should treat prompts and schemas as a unified instruction surface and validate placement empirically for each target model, keeping in mind that engineering the schema's output fields directly may be a more effective lever than placement alone.
\end{abstract}

\section{Introduction}

Large language models are increasingly deployed as structured predictors rather than free-form generators. In production systems, models often do not produce unconstrained text; they populate predefined JSON schemas for classification, information extraction, evaluation, and agent tool calls. Structured output has become a standard interface for building reliable LLM applications, particularly in workflows where downstream systems require predictable fields and types \citep{openai2024structured,geng2025jsonschemabench}.

When these systems fail, the default response is prompt optimization. Practitioners rewrite instructions, add examples, or use automated prompt optimization methods to improve model behavior \citep{zhou2023ape,yang2024opro}. These approaches implicitly assume that the prompt is the primary location where task instructions are interpreted. Structured-output systems challenge this assumption. A schema property can contain a description field that is conventionally treated as metadata, which describes the meaning or expected content of a field rather than serving as an explicit instruction channel. However, when these descriptions are provided to an LLM as part of a schema, they may influence model behavior beyond serving as metadata. There are currently no clear rules governing what should or should not be included in these fields. In practice, descriptions can contain almost anything, including category definitions, behavioral constraints, conditional rules, or other task guidance.

The role of schema-level instructions remains unclear in vendor provided examples. Common practice often treats system prompts as the authoritative instruction source and schema descriptions as secondary metadata. At the same time, API examples from major LLM providers demonstrate schema descriptions containing conditional logic and behavioral guidance, suggesting that schemas can influence model decisions \citep{openai2024structured}. For example, structured-output examples include descriptions such as ``Type of violation, if the content is violating guidelines. Null otherwise,'' which specifies when a field should be populated rather than merely describing its format \citep{openai2024structured}. Meanwhile, practitioners report cases where schema descriptions appear ineffective and move definitions into prompts instead \citep{pydanticai2025issue}. Together, these observations raise a fundamental question:

\begin{quote}
\emph{How does the placement of an instruction---in the prompt versus the schema---affect model behavior?}
\end{quote}

We encountered this question through a practical case. In a classification task, category definitions were placed in schema descriptions following a common structured-output pattern. One category repeatedly failed. The model assigned short keyword inputs to a category intended only for full questions. Rewriting the definition did not resolve the issue. However, moving the identical definition text from the schema into the system prompt immediately improved performance.

This observation motivated a direct measurement of instruction placement effects. Prior work has examined several related questions, but none directly isolates where an instruction is placed within an LLM API. Studies of structured generation and output constraints have investigated whether formatting requirements affect model performance. Some report performance degradation under format restrictions \citep{tam2024letmespeak}, while others argue that these effects may instead arise from prompting artifacts \citep{dottxt2024saywhatyoumean}. Benchmarking work has also measured structured-output reliability directly, without examining whether the placement of instructions within the API contributes to model behavior \citep{geng2025jsonschemabench}. A separate line of work shows that semantically equivalent instructions can produce different outcomes depending on how they are formulated, but focuses primarily on variations within the prompt itself \citep{sclar2024quantifying}. Research on context-position effects similarly demonstrates that where information appears can affect model behavior, but has focused on retrieved or in-context information rather than instructions embedded in structured-output schemas \citep{liu2024lostinmiddle}.

To answer the open question, in this work, we measured how LLMs interpret schema-level instructions relative to prompt-level instructions. We used a controlled single-field classification task with nonce category labels, allowing us to isolate instruction placement effects without relying on pretrained associations with category names. Across ten model configurations from two vendors, we moved identical definitions between the system prompt, user prompt, and schema description, and introduced conflicting instructions to directly test whether schema content can override prompt instructions.

Our contributions are:

\begin{enumerate}
  \item \textbf{A controlled instruction placement study.} We measured the effect of moving identical definition text between system prompts, user prompts, and schema descriptions across multiple LLM configurations.
  \item \textbf{A conflict-based measurement of schema authority.} We introduced cases where prompts and schemas provide contradictory definitions. These experiments showed that schema descriptions are not passive metadata: some models substantially incorporated schema instructions even when they conflicted with prompts.
  \item \textbf{A model-dependent view of schema reliance.} We showed that different models assign different levels of influence to schema instructions. Some under-utilize schema descriptions, while others allow schema instructions to override prompt instructions.
  \item \textbf{A schema design intervention.} We showed that adding a required intermediate field before the prediction field can substantially improve schema-only classification performance for all tested models with performance headroom, including a model already equipped with reasoning abilities, suggesting that schema structure itself affects instruction execution.
\end{enumerate}

\section{Method}

\subsection{Task}

We designed a single-field classification task based on the \emph{claim\_order} taxonomy, which requires the model to classify short workplace messages according to their argumentative structure. Each message belongs to one of four categories: conclusion-then-reason, reason-then-conclusion, conclusion with no reason, or reason with the conclusion left implicit. The categories were represented using nonce labels (\texttt{tomil}, \texttt{varek}, \texttt{selmo}, \texttt{drava}) to prevent label semantics from providing classification cues.

The evaluation set contained 48 items, with 12 examples per category and manually verified ground truth. For example, ``We should cancel the launch. The servers can't handle the load.'' corresponds to the conclusion-then-reason category.

Nonce labels are essential for isolating instruction placement effects. If models could infer category meaning from label names, performance would not reflect whether definitions were available or where they were placed. We therefore included a \texttt{labels\_only} control condition containing no category definitions. Performance in this condition should approach the 25\% chance level expected from four balanced categories.

\subsection{Experimental Conditions}

All conditions used the same structured-output schema and identical category definitions. We varied only the location and structure through which the definitions were provided (Table~\ref{tab:conditions}).

\begin{table}[h]
  \centering
  \caption{Experimental conditions and where category definitions live in each.}
  \label{tab:conditions}
  \begin{tabular}{ll}
    \toprule
    Condition & Definition location \\
    \midrule
    \texttt{labels\_only} & No definitions (control condition) \\
    \texttt{schema\_only} & Schema description fields \\
    \texttt{system\_prompt\_only} & System prompt \\
    \texttt{user\_prompt} & User prompt \\
    \texttt{system\_prompt\_start/end} & System prompt, position varied via filler text \\
    \texttt{system\_prompt\_json\_block} & System prompt, definitions as a JSON code block \\
    \texttt{conflicting} & Correct in system prompt, incorrect in schema \\
    \texttt{schema\_with\_cot} & Schema, plus a required reasoning field before the label field \\
    \bottomrule
  \end{tabular}
\end{table}

The incorrect schema definitions in the \texttt{conflicting} condition were constructed by permuting the correct definitions across categories: each category's schema description was replaced with another category's correct definition, so the schema text remained fluent and internally coherent, just assigned to the wrong category, rather than being random or obviously nonsensical text. The \texttt{conflicting} condition was evaluated on models with sufficient performance headroom in the corresponding consistent condition. Models already achieving near-ceiling accuracy were excluded from this comparison because their error rate limits the interpretability of additional performance changes. If schema descriptions function only as formatting metadata, incorrect schema definitions should not affect predictions relative to the \texttt{system\_prompt\_only} condition. Performance degradation under conflict indicates that the model incorporates schema-level instructions when making predictions.

\subsection{Models and measurement}

We evaluated ten model configurations from two vendors: GPT-4.1, GPT-5.4, GPT-5.5, Claude Haiku 4.5, Claude Sonnet 4.6, and Claude Opus 4.8, with separate configurations for available reasoning levels. For reasoning-enabled models, we evaluated both no-reasoning and medium-reasoning settings where supported.

Each condition was evaluated five times with reshuffled item order over the 48-item evaluation set (12 items per category). We reported mean accuracy across these five repeats and quantified uncertainty from two sources: repeat standard error (SE) and binomial SE. Repeat SE was computed as the standard deviation across the five runs divided by $\sqrt{5}$. It measures run-to-run variability under our evaluation procedure and serves as the primary indicator of reproducibility for condition comparisons. Binomial SE was computed as $\sqrt{p(1-p)/n}$ with $n=48$. It reflects uncertainty associated with estimating accuracy from a compact evaluation set and is reported for completeness, but not used for significance testing. The evaluation set size was intentionally compact to enable repeated evaluation across a broad set of model configurations while maintaining controlled comparisons across conditions; therefore, we focused on within-task performance differences rather than absolute accuracy estimates. Repeat SE was at or below 1.5 percentage points for all conditions discussed in our analyses. The only exception was the \texttt{labels\_only} control for GPT-5.5, which showed repeat SE of 3.93 points. Because this control is used only to verify nonce-label neutrality rather than to compare instruction conditions, this variability does not affect the conclusions of the study.

\section{Results}

\subsection{Instruction Placement Effects}
\label{sec:placement}

We first evaluated whether identical category definitions produced different classification performance depending on where they were provided. The primary comparison conditions are reported in Table~\ref{tab:table1}. Additional analyses, including binomial standard errors and control conditions such as JSON-formatted system prompts and system prompt position variants, are provided in Appendix~\ref{app:a} and Appendix~\ref{app:b}.

The \texttt{labels\_only} control confirmed that nonce labels do not provide semantic cues. Most models remained near the 25\% chance level (20.83--33.33\%). GPT-4.1 achieved higher apparent accuracy (45.83\% averaged over five repeats). A representative single-run confusion matrix from this condition (43.75\% on that run, close to the five-run mean) showed a degenerate prediction strategy: the model never predicted \texttt{tomil} or \texttt{varek}, collapsing all predictions onto \texttt{selmo} and \texttt{drava}. Above-chance accuracy came from this collapse happening to align with two of the four true categories in this test set, not from reading label semantics. We therefore did not interpret this result as evidence of label understanding.

The results showed that schema descriptions function as an instruction channel, but they are not uniformly equivalent to prompt-based instructions. Compared with the \texttt{labels\_only} condition, providing definitions through schema descriptions substantially improved classification accuracy across models, demonstrating that models can utilize schema-embedded instructions. However, for GPT-4.1 and GPT-5.4 without reasoning, moving definitions from the schema into the system prompt improved accuracy by 12.50 and 11.25 percentage points, respectively (GPT-4.1: schema 77.08\% vs.\ system prompt 89.58\%; GPT-5.4: schema 80.00\% vs.\ system prompt 91.25\%). These models performed substantially better when definitions were provided through the system prompt, suggesting that they rely more heavily on system-level instructions than on schema descriptions.

In contrast, several models showed comparable performance between schema and system prompt placements when instructions were consistent. Claude Haiku 4.5 achieved similar but relatively low performance under both placements (schema 53.75\% vs.\ system prompt 52.50\%). Claude Sonnet 4.6 also showed minimal differences between placements (no reasoning: schema 82.92\% vs.\ system prompt 83.75\%; medium reasoning: schema 81.67\% vs.\ system prompt 83.33\%). Claude Opus 4.8 and GPT-5.5 reached ceiling performance under both placements. These results indicated that schema descriptions can serve as an effective instruction location for some models, but their reliability varies substantially across model families.

Overall, system prompt placement provides the most reliable default for storing classification definitions. However, schema descriptions should not be treated as mere formatting metadata: providing definitions through the schema substantially improves performance over the labels-only condition, demonstrating that models can utilize schema-embedded instructions. At the same time, the degree of schema utilization varies across models, and schema descriptions should be empirically validated when used as an instruction channel.

\begin{table}[h]
  \centering
  \caption{Core instruction placement experiment: classification accuracy (\%) with repeat SE in parentheses. \textbf{Bold} marks the row maximum (ties all bolded).}
  \label{tab:table1}
  \small
  \begin{tabular}{llrrrr}
    \toprule
    Model & Reasoning & Labels Only & Schema Only & System Prompt & User Prompt \\
    \midrule
    GPT-4.1            & none   & 45.83 (0.93) & 77.08 (0.66)          & \textbf{89.58} (0.93) & 74.58 (0.42) \\
    GPT-5.4            & none   & 27.50 (0.42) & 80.00 (0.51)          & \textbf{91.25} (0.42) & 68.33 (1.21) \\
    GPT-5.4            & medium & 25.00 (0.66) & \textbf{100.00} (0.00) & \textbf{100.00} (0.00) & \textbf{100.00} (0.00) \\
    GPT-5.5            & none   & 27.92 (3.93) & \textbf{100.00} (0.00) & \textbf{100.00} (0.00) & \textbf{100.00} (0.00) \\
    GPT-5.5            & medium & 20.83 (2.19) & \textbf{100.00} (0.00) & \textbf{100.00} (0.00) & \textbf{100.00} (0.00) \\
    Claude Haiku 4.5   & none   & 25.00 (0.00) & 53.75 (0.42)          & 52.50 (0.42)          & \textbf{58.75} (0.42) \\
    Claude Sonnet 4.6  & none   & 31.67 (1.02) & 82.92 (0.78)          & \textbf{83.75} (0.42) & 78.33 (0.51) \\
    Claude Sonnet 4.6  & medium & 33.33 (1.47) & 81.67 (0.42)          & \textbf{83.33} (0.66) & 78.75 (0.78) \\
    Claude Opus 4.8    & none   & 27.08 (1.32) & \textbf{100.00} (0.00) & \textbf{100.00} (0.00) & 96.25 (0.42) \\
    Claude Opus 4.8    & medium & 24.58 (0.42) & 97.92 (0.93)          & \textbf{98.75} (0.51) & 97.92 (0.00) \\
    \bottomrule
  \end{tabular}
\end{table}

\subsection{The Conflict Probe}
\label{sec:conflict}

The placement experiment established that models can utilize schema descriptions, but it did not reveal whether schema content remains passive when it conflicts with prompt instructions. We therefore introduced a conflict condition in which the system prompt contained the correct category definitions while the schema description contained intentionally incorrect definitions. If schema descriptions function only as formatting metadata, performance should remain comparable to the system-prompt-only condition. Instead, we observed substantial and model-dependent performance changes (Table~\ref{tab:table2}).

The conflict condition revealed that schema descriptions influence model predictions even when the prompt contains the correct instructions. Claude Haiku 4.5 showed the strongest schema sensitivity: accuracy dropped from 52.50\% to 7.08\% ($-45.42$ points), indicating that the incorrect schema definitions largely determine the model's classification behavior. GPT-5.5 also exhibited strong conflict sensitivity, with accuracy decreasing from 100\% to 73.33\% in the no-reasoning setting and to 75.83\% with medium reasoning ($-26.67$ and $-24.17$ points, respectively). Claude Sonnet 4.6 similarly showed substantial degradation under conflicting schemas ($-17.92$ and $-19.58$ points).

In contrast, GPT-4.1 and GPT-5.4 without reasoning showed minimal sensitivity to conflicting schema definitions. GPT-4.1 decreased by only 2.91 points, while GPT-5.4 slightly improved by 2.50 points. GPT-5.4 with medium reasoning and both Claude Opus 4.8 configurations remained near ceiling performance, limiting observable effects. These results indicated that schema descriptions do not have a fixed position in an instruction hierarchy. Instead, models differ substantially in how strongly they incorporate schema-level instructions relative to prompt-level instructions.

\begin{table}[h]
  \centering
  \caption{Model performance change when conflicting information is shown in the schema. $\Delta$ is Conflicting minus Consistent (System-Prompt Only).}
  \label{tab:table2}
  \begin{tabular}{llrrr}
    \toprule
    Model & Reasoning & System-Prompt Only & Conflicting Schema & $\Delta$ Conflict \\
    \midrule
    GPT-4.1            & none   & \textbf{89.58} (0.93) & 86.67 (1.06)          & $-2.9$ \\
    GPT-5.4            & none   & 91.25 (0.42)          & \textbf{93.75} (0.00) & $+2.5$ \\
    GPT-5.4            & medium & \textbf{100.00} (0.00) & \textbf{100.00} (0.00) & $0.0$ \\
    GPT-5.5            & none   & \textbf{100.00} (0.00) & 73.33 (2.98)          & $\mathbf{-26.7}$ \\
    GPT-5.5            & medium & \textbf{100.00} (0.00) & 75.83 (3.27)          & $\mathbf{-24.2}$ \\
    Claude Haiku 4.5   & none   & \textbf{52.50} (0.42) & 7.08 (0.51)           & $\mathbf{-45.4}$ \\
    Claude Sonnet 4.6  & none   & \textbf{83.75} (0.42) & 65.83 (0.51)          & $\mathbf{-17.9}$ \\
    Claude Sonnet 4.6  & medium & \textbf{83.33} (0.66) & 63.75 (0.83)          & $\mathbf{-19.6}$ \\
    Claude Opus 4.8    & none   & \textbf{100.00} (0.00) & 93.33 (1.53)          & $\mathbf{-6.7}$ \\
    Claude Opus 4.8    & medium & \textbf{98.75} (0.51) & 94.17 (0.78)          & $\mathbf{-4.6}$ \\
    \bottomrule
  \end{tabular}
\end{table}

\subsection{Schema Structure Intervention}
\label{sec:cot}

One potential explanation for models underusing schema-embedded definitions is a difference in how they process information across input channels. Prompt text may receive more direct instruction-following attention, while schema fields may be treated primarily as structural constraints to satisfy. If so, simply providing definitions in the schema may not be sufficient to elicit the same degree of engagement as providing them in the prompt. We therefore tested whether requiring an intermediate reasoning step within the schema could increase the model's use of schema-embedded definitions. The reasoning field was ordered before the label field, requiring the model to generate free-text reasoning as part of the schema output before producing the final label. This does not alter the position of classification instruction; instead, it introduces an additional generation step in which the model must engage with the schema-constrained task before committing to a label.

Table~\ref{tab:table3} compares the original \texttt{schema\_only} condition with this augmented schema for models that had headroom for improvement. The category definitions remained unchanged and were provided only through the schema description. Adding the intermediate reasoning field substantially improved performance for models with limited schema utilization. GPT-4.1 improved from 77.08\% to 97.92\% ($+20.83$ points), GPT-5.4 without reasoning improved from 80.00\% to 100.00\% ($+20.00$ points), and Claude Haiku 4.5 improved from 53.75\% to 77.50\% ($+23.75$ points). Claude Sonnet 4.6 similarly improved by 14.58 and 17.50 points for the no-reasoning and medium-reasoning settings, respectively.

The intervention brought every reported model except Claude Haiku 4.5 close to ceiling performance, and exceeded system-prompt-only performance for every model where a direct comparison was available. Notably, even Claude Sonnet 4.6 with medium reasoning effort showed a substantial improvement, increasing from 81.67\% to 99.17\%. These results suggest that the benefit comes not simply from enabling additional reasoning computation, but from how the schema structure organizes and exposes the task-relevant information.

\begin{table}[h]
  \centering
  \caption{Impact of an intermediate CoT schema field on classification accuracy. Restricted to configurations with measurable headroom under \texttt{schema\_only} and available intervention data.}
  \label{tab:table3}
  \begin{tabular}{llrrr}
    \toprule
    Model & Reasoning & Schema Only & Schema + Intermediate Field & $\Delta$ \\
    \midrule
    GPT-4.1            & none   & 77.08 (0.66) & \textbf{97.92} (0.00) & $\mathbf{+20.8}$ \\
    GPT-5.4            & none   & 80.00 (0.51) & \textbf{100.00} (0.00) & $\mathbf{+20.0}$ \\
    Claude Haiku 4.5   & none   & 53.75 (0.42) & \textbf{77.50} (1.02) & $\mathbf{+23.8}$ \\
    Claude Sonnet 4.6  & none   & 82.92 (0.78) & \textbf{97.50} (0.42) & $\mathbf{+14.6}$ \\
    Claude Sonnet 4.6  & medium & 81.67 (0.42) & \textbf{99.17} (0.51) & $\mathbf{+17.5}$ \\
    \bottomrule
  \end{tabular}
\end{table}

GPT-5.4 (medium), GPT-5.5 (none/medium), and Claude Opus 4.8 (both reasoning settings) reached 100.00\% or near-100.00\% under \texttt{schema\_only} already (Table~\ref{tab:table1}) and were omitted here, since there was no room for the intervention to show an effect regardless of whether it was run.

\section{Discussion}

\paragraph{Structured outputs introduce a second instruction interface.}
Our results de-mystify the common abstraction that structured outputs separate \emph{behavioral instructions} (prompts) from \emph{formatting constraints} (schemas), rather than simply confirming or denying it. Across ten model configurations, schema descriptions consistently affected behavior: even models that underused schema descriptions in the placement experiment still showed measurable schema influence in the conflict condition. The key difference was not whether models read schema descriptions, but how strongly they weighted them relative to other instruction sources. This reframes structured-output design as an instruction-placement problem rather than solely a serialization problem. Further, the variance in prompt versus schema weighting across models suggests that prompt placement is an empirical question, and developers cannot infer instruction priority from the interface alone.

\paragraph{Instruction consistency matters more than instruction location.}
A practical implication is that prompts and schemas should be treated as a unified instruction surface. The highest-risk failure mode we observed was not a schema description being ignored, but a schema description silently overriding a correct prompt. GPT-5.5 provides a representative example: it showed no measurable difference between prompt and schema placement when instructions were consistent, but accuracy dropped by 24--27 points when the schema contained conflicting definitions, regardless of reasoning setting.

Therefore, definitions and behavioral constraints should have a single source of truth and be rendered into the instruction locations that are most reliable for a given model. Maintaining separate prompt and schema copies creates exactly the type of inconsistency that our conflict probe revealed as harmful.

\paragraph{Instruction placement should be measured before prompt optimization.}
Current prompt optimization workflows typically assume that the prompt is the primary control surface. Our results suggest that this assumption is incomplete for structured-output applications. On models that heavily weight schema descriptions, improving the prompt alone may not resolve failures caused by schema-level instructions. Conversely, on models that underweight schemas, adding more information to schema field descriptions may have limited impact.

A lightweight diagnostic can identify the relevant instruction hierarchy before optimization begins. In our setting, a clean placement condition and a conflicting condition required fewer than 100 API evaluations to characterize whether a model discounted or incorporated schema instructions. Such measurement can guide where engineering effort should be invested.

\paragraph{Schema structure is itself an intervention surface.}
Beyond instruction placement, our intermediate-reasoning-field experiment suggests that the schema structure itself affects how models apply schema-provided information. Adding a required intermediate reasoning field improved performance for all models with available performance headroom, including cases where native reasoning settings did not provide comparable gains. This indicates that structured-output design is not only a constraint on model outputs but also a mechanism for organizing model computation.

The broader implication is that reliable structured-output systems require both appropriate instruction content and appropriate schema architecture.

\paragraph{Measuring instruction placement remains a moving target.}
An additional observation from developing this benchmark is that placement effects are only measurable when the task requires the model to rely on external definitions. Many intuitive classification tasks produced ceiling performance regardless of instruction location because frontier models already inferred the intended categories. The measurable regime is therefore narrower than expected and shifts as models improve.

Notably, the largest placement effects appeared in configurations commonly used for cost-efficient deployment: capable models with reasoning disabled. These settings may receive less attention in benchmark evaluations but remain important for production-scale labeling and extraction pipelines.

\paragraph{Practical and broader impact.}
Beyond the specific findings, this work has three direct, reusable applications for practitioners building structured-output classification systems. First, it replaces guesswork with a cheap, reusable diagnostic: a clean and a conflicting condition (under 100 API calls per model configuration, Section~4) tell a practitioner which instruction channel their specific model actually follows, rather than requiring trial-and-error prompt engineering informed only by anecdote. Second, it provides controlled empirical evidence for a question that has so far only been debated anecdotally on developer forums and in vendor documentation, where reports of schema descriptions being ignored \citep{pydanticai2025issue} and vendor examples that embed conditional instructions in descriptions \citep{openai2024structured} point in opposite directions; our results show both are correct for different models, which settles the disagreement rather than adding another anecdote to it. Third, the intermediate-reasoning-field intervention (Section~3.3) is not merely a diagnosed mechanism but a directly deployable fix: adding a required reasoning field before the label field in a schema is a one-line schema change practitioners can apply immediately, without further experimentation, wherever they have already confirmed their model under-weights the schema channel.

This last point carries a caution the paper itself argues for: the intervention should be verified per-model rather than applied blindly, since its benefit is unmeasurable in models already at ceiling (Table~\ref{tab:table3}). Treating any single recommendation in this paper, including the intervention, as universal rather than something to measure would repeat the exact failure mode the paper diagnoses in current prompt-engineering practice. We do not identify a negative societal impact specific to this contribution: the work does not release a model, a scraped dataset, or a capability with a plausible dual-use path, and its expected effect is more reliable, auditable behavior in structured-output pipelines that already exist.

\section{Limitations}

This study has several limitations. First, our evaluation focuses on a single-field classification task with a compact evaluation set of 48 examples (12 examples per category) and nonce category labels. The small sample size limits the precision of absolute accuracy estimates and makes the benchmark unsuitable for estimating general task performance. We chose this compact design to enable repeated evaluations across ten model configurations while maintaining strict control over instruction content and placement. Accordingly, our conclusions focus on relative performance differences between controlled conditions rather than absolute accuracy estimates. While the design isolates instruction placement effects, it does not establish how schema influence generalizes to all structured-output tasks, such as multi-field extraction, tool calling, or complex agent workflows. Our additional multi-field experiments did not reveal reliable placement effects in either direction, and we therefore make no claims about whether schema co-location helps or hurts when many fields are predicted simultaneously.

Second, our conflict condition represented a maximal disagreement: the schema definitions were fully permuted relative to the correct prompt definitions. Real-world failures may arise from weaker forms of inconsistency, such as stale descriptions, incomplete paraphrases, or partially updated documentation. Future work should characterize the dose-response relationship between degrees of schema drift and model behavior.

Third, the proposed schema-weight interpretation is an empirical description of observed behavior rather than a mechanistic explanation. Without access to model internals, we cannot determine whether schema influence arises from attention patterns, instruction hierarchy training, or other factors. We use ``schema weight'' as a practical behavioral characterization: a reproducible measurement of how strongly a model follows schema-level instructions relative to other instruction sources.

Finally, model behavior changes over time. The measurements in this work characterize ten configurations at a particular point in development. As models and APIs evolve, instruction placement should be treated as a property to measure rather than a fixed assumption.

\section{Conclusion}

Structured outputs are often treated as a separation between instructions and formatting: prompts define behavior, while schemas define output structure. Our results show that this abstraction is incomplete. Schema descriptions function as an additional instruction channel, but models differ substantially in how they weight this channel relative to prompts. Some models underutilize schema descriptions, leading to lower performance when definitions are moved from the prompt into the schema, while others strongly incorporate schema instructions and can be affected by incorrect schema content even when the prompt is correct.

These findings suggest a different design principle for structured-output systems: instruction placement is a model-dependent property that should be measured rather than assumed. Across the models tested, the system prompt provided a reliable default for task-critical instructions, while schema descriptions varied substantially in how effectively they were used. Developers should therefore treat prompts and schemas as a unified instruction surface, maintain definitions as a single source of truth, and avoid duplicated or inconsistent instructions across locations. A lightweight placement and conflict evaluation can reveal which instruction channels a target model relies on before additional effort is spent on prompt optimization.

Finally, our schema structure intervention demonstrates that the output schema itself can shape model behavior. Adding an intermediate required reasoning field improved performance for models with limited schema utilization, including cases where native reasoning settings did not provide comparable gains. Structured outputs are therefore not only a mechanism for constraining model outputs; they are also a design interface for influencing model behavior.

As models and APIs continue to evolve, the question is no longer whether instructions belong in the prompt or the schema. Instead, reliable LLM applications require understanding how each model utilizes the available instruction channels and designing the full instruction interface accordingly.

\paragraph{Data and code availability.} We release the evaluation harness, the \emph{claim\_order} taxonomy, the 48-item test set, and all condition configurations used in Sections~2--3, so that the placement and conflict diagnostics can be rerun as new model versions are released: \url{https://github.com/alina-lin-phd/prompt-placement-eval}.

\bibliographystyle{plainnat}
\bibliography{references}

\appendix

\section{Prompt Formatting and Position Controls}
\label{app:a}

These conditions held definitions in the system prompt throughout and varied only their position or surrounding syntax: \texttt{system\_prompt\_start}/\texttt{system\_prompt\_end} moved the definition block to the beginning or end of the prompt via filler text; \texttt{json\_block} wrapped the same definitions in a JSON code block rather than prose. All values were drawn from the same runs as Table~\ref{tab:table1} and reported as accuracy\% (repeat SE\%). \textbf{Bold} marks the row maximum.

\begin{table}[h]
  \centering
  \caption{Prompt formatting and position controls (\S\ref{sec:placement}).}
  \label{tab:appa}
  \footnotesize
  \begin{tabular}{llrrrr}
    \toprule
    Model & Reasoning & System Prompt & Sys.\ Prompt Start & Sys.\ Prompt End & JSON Block \\
    \midrule
    GPT-4.1            & none   & \textbf{89.58} (0.93) & 83.75 (0.78)          & 88.75 (0.51)          & 82.92 (1.21) \\
    GPT-5.4            & none   & 91.25 (0.42)          & \textbf{93.75} (0.00) & 90.00 (0.42)          & 78.75 (0.78) \\
    GPT-5.4            & medium & \textbf{100.00} (0.00) & \textbf{100.00} (0.00) & \textbf{100.00} (0.00) & \textbf{100.00} (0.00) \\
    GPT-5.5            & none   & \textbf{100.00} (0.00) & \textbf{100.00} (0.00) & \textbf{100.00} (0.00) & \textbf{100.00} (0.00) \\
    GPT-5.5            & medium & \textbf{100.00} (0.00) & \textbf{100.00} (0.00) & \textbf{100.00} (0.00) & \textbf{100.00} (0.00) \\
    Claude Haiku 4.5   & none   & 52.50 (0.42)          & 52.08 (0.00)          & 55.00 (0.51)          & \textbf{55.83} (0.42) \\
    Claude Sonnet 4.6  & none   & \textbf{83.75} (0.42) & 78.75 (1.21)          & 82.08 (1.06)          & 82.50 (1.06) \\
    Claude Sonnet 4.6  & medium & \textbf{83.33} (0.66) & 82.08 (0.51)          & 82.08 (0.83)          & 82.50 (0.83) \\
    Claude Opus 4.8    & none   & \textbf{100.00} (0.00) & \textbf{100.00} (0.00) & \textbf{100.00} (0.00) & 97.50 (0.42) \\
    Claude Opus 4.8    & medium & 98.75 (0.51)          & \textbf{99.17} (0.51) & 98.33 (0.78)          & 97.92 (0.00) \\
    \bottomrule
  \end{tabular}
\end{table}

The system prompt matched or beat every position/formatting variant for every model except GPT-5.4 (none), where moving definitions to the very start of the prompt scored slightly higher (prompt start 93.75\% vs.\ system prompt 91.25\%), and Claude Haiku 4.5, where a JSON-formatted block scored slightly higher (JSON block 55.83\% vs.\ system prompt 52.50\%). Both gaps are within 3 points and do not change the placement conclusions in Section~\ref{sec:placement}: position within the prompt and prose-vs-JSON formatting are minor, model-specific effects relative to the prompt-vs-schema boundary that is the paper's focus.

\section{Binomial SE Reference}
\label{app:b}

Binomial SE (\%), computed as $\sqrt{p(1-p)/n}$ with $n=48$, for every cell reported in Tables~\ref{tab:table1}--\ref{tab:table2} and Table~\ref{tab:appa}, plus the \texttt{schema\_only} baseline for all ten configurations (Table~\ref{tab:table3} in the main text reports only the five configurations with measurable headroom; the remaining five are included here for completeness). As discussed in Section~2.3, we do not use binomial SE as a significance criterion; it is reported here for readers who want single-run sampling variance alongside the point estimates. Note that as a normal approximation it understates uncertainty near 0\% or 100\% accuracy, where many cells below sit.

\begin{table}[h]
  \centering
  \caption{Binomial SE for Tables~\ref{tab:table1}--\ref{tab:table2} conditions.}
  \footnotesize
  \begin{tabular}{llrrrrr}
    \toprule
    Model & Reasoning & Labels Only & Schema Only & System Prompt & User Prompt & Conflicting \\
    \midrule
    GPT-4.1            & none   & 7.19 & 6.07 & 4.41 & 6.28 & 4.91 \\
    GPT-5.4            & none   & 6.44 & 5.77 & 4.08 & 6.71 & 3.49 \\
    GPT-5.4            & medium & 6.25 & 0.00 & 0.00 & 0.00 & 0.00 \\
    GPT-5.5            & none   & 6.47 & 0.00 & 0.00 & 0.00 & 6.38 \\
    GPT-5.5            & medium & 5.86 & 0.00 & 0.00 & 0.00 & 6.18 \\
    Claude Haiku 4.5   & none   & 6.25 & 7.20 & 7.21 & 7.11 & 3.70 \\
    Claude Sonnet 4.6  & none   & 6.71 & 5.43 & 5.32 & 5.95 & 6.85 \\
    Claude Sonnet 4.6  & medium & 6.80 & 5.58 & 5.38 & 5.90 & 6.94 \\
    Claude Opus 4.8    & none   & 6.41 & 0.00 & 0.00 & 2.74 & 3.60 \\
    Claude Opus 4.8    & medium & 6.21 & 2.06 & 1.60 & 2.06 & 3.38 \\
    \bottomrule
  \end{tabular}
\end{table}

\begin{table}[h]
  \centering
  \caption{Binomial SE for the full Table~\ref{tab:table3} grid (schema structure intervention).}
  \begin{tabular}{llr}
    \toprule
    Model & Reasoning & Schema Only \\
    \midrule
    GPT-4.1            & none   & 6.07 \\
    GPT-5.4            & none   & 5.77 \\
    GPT-5.4            & medium & 0.00 \\
    GPT-5.5            & none   & 0.00 \\
    GPT-5.5            & medium & 0.00 \\
    Claude Haiku 4.5   & none   & 7.20 \\
    Claude Sonnet 4.6  & none   & 5.43 \\
    Claude Sonnet 4.6  & medium & 5.58 \\
    Claude Opus 4.8    & none   & 0.00 \\
    Claude Opus 4.8    & medium & 2.06 \\
    \bottomrule
  \end{tabular}
\end{table}

\begin{table}[h]
  \centering
  \caption{Binomial SE for Table~\ref{tab:appa} (prompt formatting and position controls).}
  \footnotesize
  \begin{tabular}{llrrrr}
    \toprule
    Model & Reasoning & System Prompt & Sys.\ Prompt Start & Sys.\ Prompt End & JSON Block \\
    \midrule
    GPT-4.1            & none   & 4.41 & 5.32 & 4.56 & 5.43 \\
    GPT-5.4            & none   & 4.08 & 3.49 & 4.33 & 5.90 \\
    GPT-5.4            & medium & 0.00 & 0.00 & 0.00 & 0.00 \\
    GPT-5.5            & none   & 0.00 & 0.00 & 0.00 & 0.00 \\
    GPT-5.5            & medium & 0.00 & 0.00 & 0.00 & 0.00 \\
    Claude Haiku 4.5   & none   & 7.21 & 7.21 & 7.18 & 7.17 \\
    Claude Sonnet 4.6  & none   & 5.32 & 5.90 & 5.54 & 5.48 \\
    Claude Sonnet 4.6  & medium & 5.38 & 5.54 & 5.54 & 5.48 \\
    Claude Opus 4.8    & none   & 0.00 & 0.00 & 0.00 & 2.25 \\
    Claude Opus 4.8    & medium & 1.60 & 1.31 & 1.85 & 2.06 \\
    \bottomrule
  \end{tabular}
\end{table}

\end{document}